\documentclass[letterpaper, 10 pt, conference]{ieeeconf}  

\IEEEoverridecommandlockouts

\let\labelindent\relax
\usepackage{algorithm}
\usepackage{algorithmicx}
\usepackage{algpseudocode}
\usepackage{times}
\usepackage{multicol}
\usepackage[bookmarks=true]{hyperref}
\usepackage{graphicx}
\usepackage{amsmath,amssymb,amsfonts,mathabx,amsbsy,mathtools,etoolbox,amsthm}

\usepackage{cleveref}

\newtheorem{assumption}{Assumption}[section]

\usepackage{bm} 
\usepackage{hyperref} 
\usepackage{enumitem}

\usepackage{booktabs,tabularx,colortbl,multirow,multicol,array,makecell}
 \usepackage{comment}

 \usepackage{cite}
            
\makeatletter
\DeclareRobustCommand\onedot{\futurelet\@let@token\@onedot}
\def\@onedot{\ifx\@let@token.\else.\null\fi\xspace}

\makeatother

\DeclareMathOperator*{\argmin}{arg\,min}

\makeatletter
\def\BState{\State\hskip-\ALG@thistlm}
\makeatother

\newtheorem{proposition}{Proposition}

\makeatletter
\renewcommand{\paragraph}{%
  \@startsection{paragraph}{4}%
  {\z@}{0ex \@plus 0ex \@minus 0ex}{-1em}%
  {\hskip\parindent\normalfont\normalsize\bfseries}%
}
\makeatother

\crefname{algorithm}{Alg.}{Algs.}
\Crefname{algocf}{Algorithm}{Algorithms}
\crefname{section}{Sec.}{Secs.}
\Crefname{section}{Section}{Sections}
\crefname{table}{Tab.}{Tabs.}
\Crefname{table}{Table}{Tables}
\crefname{figure}{Fig.}{Fig.}
\Crefname{figure}{Figure}{Figure}

\let\oldnl\nl
\newcommand{\nonl}{\renewcommand{\nl}{\let\nl\oldnl}}%
\makeatother

\begin{document}
\title{\LARGE \bf
Risk-Aware Non-Myopic Motion Planner for Large-Scale Robotic Swarm Using CVaR Constraints
}
\author{Xuru Yang$^{1}$, Yunze Hu$^{1}$, Han Gao$^{1}$, 
 Kang Ding$^{1}$, Zhaoyang Li$^{2}$, Pingping Zhu$^{3}$, Ying Sun$^{4}$ and Chang Liu$^{1}$
\thanks{*This work was sponsored by Beijing Nova Program (20220484056) and the National Natural Science Foundation of China (62203018).}
\thanks{$^1$ Xuru Yang, Yunze Hu, Han Gao, Kang Ding, and Chang Liu are with the Department of Advanced Manufacturing and Robotics, College of Engineering, Peking University, Beijing 100871, China (xuru.yang@stu.pku.edu.cn; hu\_yun\_ze@stu.pku.edu.cn; hangaocoe@pku.edu.cn; kangding@stu.pku.edu.cn; changliucoe@pku.edu.cn). All correspondences should be sent to Chang Liu.}
\thanks{$^2$Zhaoyang Li is with the Department of Automation, Tsinghua University, Beijing 100080, China (lizhaoya21@mails.tsinghua.edu.cn)}
\thanks{$^3$Pingping Zhu is with the Department of Computer Sciences and Electrical Engineering (CSEE), 
Marshall University, Huntington, WV 25755, USA (zhup@marshall.edu).}
\thanks{$^4$Ying Sun is with the School of Electrical Engineering and Computer Science, The Pennsylvania State University, State College, PA 16802, USA (ybs5190@psu.edu).}}

\maketitle
\thispagestyle{empty}
\pagestyle{empty}

\begin{abstract}
Swarm robotics has garnered significant attention due to its ability to accomplish elaborate and synchronized tasks. 
Existing methodologies for motion planning of swarm robotic systems mainly encounter difficulties in scalability and safety guarantee. 
To address these limitations, we propose a Risk-aware swarm mOtion planner using conditional ValuE-at-Risk (ROVER) that systematically navigates large-scale swarms through cluttered environments while ensuring safety.
ROVER formulates a finite-time model predictive control (FTMPC) problem predicated upon the macroscopic state of the robot swarm represented by a Gaussian Mixture Model (GMM) and integrates conditional
value-at-risk (CVaR) to ensure collision avoidance. 
The key component of ROVER is imposing a CVaR constraint on the distribution of the Signed Distance Function between the swarm GMM and obstacles in the FTMPC to enforce collision avoidance.
Utilizing the analytical expression of CVaR of a GMM derived in this work, 
we develop a computationally efficient solution to solve the non-linear constrained FTMPC through sequential linear programming. Simulations and comparisons with representative benchmark approaches demonstrate the
effectiveness of ROVER in flexibility, scalability, and safety guarantee.
\end{abstract}

\IEEEpeerreviewmaketitle

\section{Introduction}
\begin{figure}[!t]
\centering
\includegraphics[width=8.5cm]{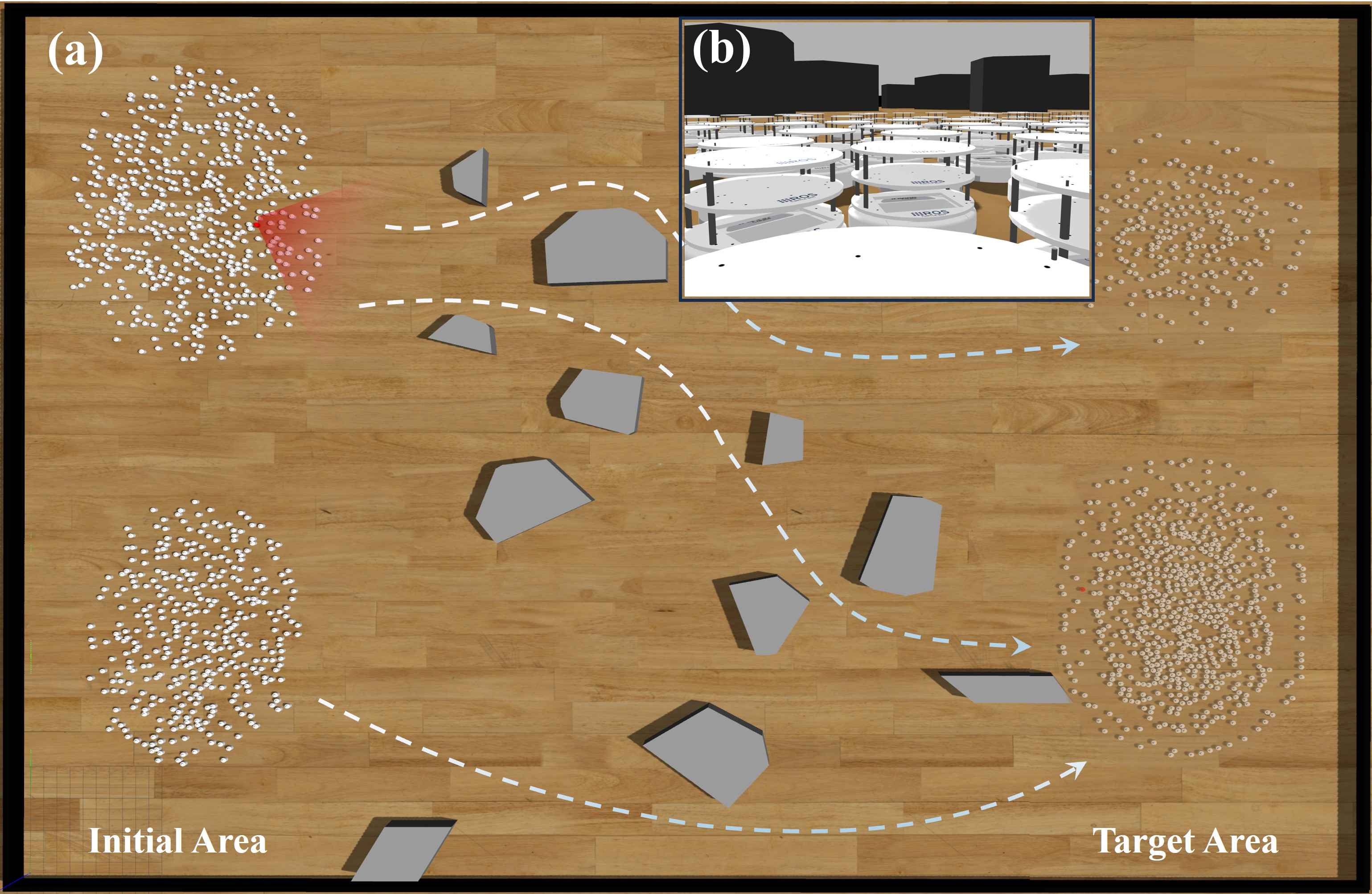}
\caption{Figure (a) illustrates the task of large-scale robotic swarm motion planning through a cluttered environment, with dashed lines indicating swarm transport trajectories and white dots denoting individual robots. Obstacles are represented as grey polygons. Figure (b) illustrates the first-person view in the red circular sector in Figure (a).}
\label{fig:coverr}
\end{figure}

Large-scale swarm robotic systems comprised of numerous autonomous and interacting robots are witnessing a surge in popularity due to their superior robustness and flexibility in applications such as target detection\cite{brambilla2013swarm}, cooperative object transport\cite{tuci2018cooperative}, and search and rescue\cite{bevacqua2015mixed}. 
In recent years, there has been a growing interest in developing motion planning techniques for large-scale swarm robots \cite{zhu2021adaptive,chen2023density}.

Present large-scale swarm motion planning approaches can be categorized into \textit{microscopic} methods and \textit{macroscopic} methods.
Microscopic methods consider agent-wise interaction and coordination and directly generate individual control inputs for each robot.
Despite the satisfying performance in small and medium-scale robot swarms \cite{soria2021predictive,soria2021distributed}, 
these methods always suffer from dramatic increases in computational overhead as the swarm size increases, hindering their applications to large-scale robotic systems.

Macroscopic approaches 
overcome the scalability issue by treating the swarm as an entity instead of focusing on individual robots. 
One representative category of macroscopic approaches is based on the mean-field theory \cite{elamvazhuthi2019mean,zheng2021transporting}, which represents the swarm state as the average of kinodynamics and cost of all robots. 
Another typical category is density control
\cite{chen2023density}, 
which models the swarm as a probability density function (PDF) and formulates the optimal control problem over the space of PDFs to transform the distribution into the target one.
These methods achieve improved scalability as the state space is independent of the swarm size.
However, these works usually assume obstacle-free environments and oversimplified agent kinematics, and cannot be implemented in real-time due to high computational demands, making them inapplicable to practical situations.

Recently, hierarchical approaches that plan the reference trajectory of the entirety of the swarm at the upper level and generate individual control for each robot to track the reference trajectory at the lower level are receiving increased attention as they integrate the advantages of both macroscopic and microscopic methods.
For example, Alonso et al. \cite{alonso2017multi} and Mao et al. \cite{mao2022optimal} proposed to generate local goals or reference trajectories for the swarm at the upper level, and developed distributed lower-level tracking controllers for each robot to track the reference trajectory. 
An evident drawback of these approaches lies in the lack of flexibility as the whole swarm is restricted to certain manually designed formations, which prevents the swarm from conducting adaptive behaviours such as splitting and merging when navigating through obstacle-cluttered environments.

To tackle this challenge, adaptive distributed optimal control (ADOC) \cite{zhu2021adaptive} proposed to model the swarm state as a Gaussian Mixture Model (GMM), and developed a macroscopic planner to navigate the GMM in a cluttered environment. 
The individual agents are then controlled by an artificial potential field (APF) method to track the GMM at the microscopic level.
ADOC is able to navigate hundreds of agents in a cluttered environment. 
However, ADOC uses myopic planning and lacks a systematic way of enforcing collision avoidance, thus leading to risky robot trajectories that are close to obstacles and cannot be easily adjusted.

A promising remedy is the conditional value-at-risk (CVaR), which has attracted significant interest as a measure of risk.
CVaR measures the expected risk that would occur beyond a certain risk level.
Compared to the traditional chance constraint risk measure, CVaR takes into account the long-tail distribution of risks and can better discern rare events.
Due to these desirable properties, CVaR has been used in motion planning of both single \cite{hakobyan2019risk} and small-scale \cite{navsalkar2023data} robotic systems.
However, there is few work adopting CVaR for large-scale robotic systems due to the high computational overhead.

To address the existing research gaps, we propose a Risk-aware mOtion planner for large-scale swarm robotics using conditional ValuE-at-Risk (ROVER).  
In particular, ROVER adopts a hierarchical motion planning framework, where the macroscopic state of the swarm is represented as a GMM, and the macroscopic planning is formulated as a finite-time model predictive control (FTMPC) problem integrating CVaR as the risk measure. 
To the best of the authors' knowledge, ROVER is the first online motion planner for large-scale swarms that analytically integrates CVaR of a GMM (GMM-CVaR) as risk constraints. 
The specific contributions of this work can be summarized as follows:
\begin{itemize}    
\item We propose to compute the distribution of Signed Distance Function (SDF) between swarm GMM and obstacles to evaluate collision between robots and obstacles. 
We further prove that the SDF distribution can be approximated as a GMM via Taylor expansion.
\item We derive an analytical expression of CVaR for GMM, which is then used as risk constraints to prevent collision between the swarm and obstacles. 
\item We present an online solution of the proposed FTMPC by Sequential Linear Programming (SLP).
Extensive simulations and comparisons with benchmark approaches verify the outstanding performance of ROVER in terms of flexibility, scalability, and risk control ability.
\end{itemize}

\section{Background and Problem formulation}
This section provides an overview of the important preliminaries followed by the problem formulation. 
For simplicity, we define a set notation $\underline{N}$ to represent the set $\{1,2,\cdots,N\}$ for any $N\in\mathbb{N}_+$.

\subsection{Wasserstein Metric}
The Wasserstein metric measures the distance between two probability distributions on a given metric space, and is widely used in the optimal transport (OT) problems \cite{chen2018optimal}. 

We first consider two Gaussian PDFs denoted by $g_1=\mathcal{N}(\boldsymbol{\mu}_1,\boldsymbol{\Sigma}_1)\in\mathcal{P}(\mathbb{R}^d)$ and $g_2=\mathcal{N}(\boldsymbol{\mu}_2,\boldsymbol{\Sigma}_2)\in\mathcal{P}(\mathbb{R}^d)$, where $\mathcal{P}(\Omega)$ denotes the set of all probability distributions defined on the sample space $\Omega$. The Wasserstein metric $W_2$ between $g_1$ and $g_2$ can be analytically calculated by \cite{chen2018optimal}:
\begin{equation}
\begin{aligned}
  W_2(g_1,g_2) &= \bigg\{\|\boldsymbol{\mu}_1-\boldsymbol{\mu}_2\|^2
\bigg.\\&\bigg.+tr\left[\boldsymbol{\Sigma}_1+\boldsymbol{\Sigma}_2- 2\left(\boldsymbol{\Sigma}_1^{1/2}\boldsymbol{\Sigma}_2\boldsymbol{\Sigma}_1^{1/2}\right)^{1/2}\right] \bigg\}^{1/2}, 
  \label{equ: wasser for gaussian}
\end{aligned}
\end{equation}
 where $tr(\cdot)$ indicates the trace operator, and $\|\cdot\|$ denotes the Euclidean norm. Furthermore, the Wasserstein metric of two GMMs can be approximated utilizing \cref{equ: wasser for gaussian} \cite{chen2018optimal}. 
 Specifically, consider two GMMs with $N_1$ and $N_2$ Gaussian components (GCs)
 denoted by 
     $\wp_1=\sum\limits_{i=1}^{N_1}\omega_1^i g_1^i,\wp_2=\sum\limits_{j=1}^{N_2}\omega_2^j g_2^j$, 
 where $g^i_1,g^j_2$ denote the $ith$ GC and the $jth$ GC of $\wp_1$ and $\wp_2$, respectively, and $\omega^i_{1},\omega^j_{2}$ denote their corresponding weights that satisfy $\sum\limits_{i=1}^{N_1}\omega_1^i=\sum\limits_{j=1}^{N_2}\omega_2^j=1$. 
 The Wasserstein metric $W_2(\wp_1,\wp_2)$ can be approximated as:
 \begin{small}
 \begin{equation}
 W_2(\wp_1,\wp_2)\triangleq \left\{\min_
{\pi\in\Pi(\boldsymbol{\omega}_1,\boldsymbol{\omega}_2)}\sum_{i=1}^{N_1}\sum_{j=1}^{N_2}[W_2(g_1^i,g_2^j)]^{2}\pi(i,j)\right\}^{1/2}.
\label{equ: GMM W2}
\end{equation}
\end{small}
where $\boldsymbol{\omega}_1=[\omega_1^1,\cdots,\omega_1^{N_1}]$ and $\boldsymbol{\omega}_2=[\omega_2^1,\cdots,\omega_2^{N_2}]$ denote the weight vector, and $\Pi(\boldsymbol{\omega}_1,\boldsymbol{\omega}_2)$ denotes the space of joint distributions between $\boldsymbol{\omega}_1$ and $\boldsymbol{\omega}_2$.
From an intuitive perspective, $\pi(i,j)$ represents the weight transported from the $ith$ GC in $\wp_1$ to the $jth$ GC in $\wp_2$.

\subsection{Signed Distance Function (SDF)}
The SDF measures the orthogonal distance between a point and the boundary of a set in a metric space with the sign indicating whether or not the point is in the interior of the set.
In particular, the SDF between a point $\boldsymbol{p}\in \mathbb{R}^2$ and an obstacle $\mathcal{O}_i \subset \mathbb{R}^2 $ can be calculated as follows, 
\begin{small}
    \begin{equation}
        sd(\boldsymbol{p},\mathcal{O}_i) = \left\{ \begin{array}{rcl}
-d(\boldsymbol{p},\partial\mathcal{O}_i), &
 \boldsymbol{p}\in \mathcal{O}_i \\ d(\boldsymbol{p},\partial\mathcal{O}_i), &  \boldsymbol{p}\notin \mathcal{O}_i 
\end{array}\right.,
\end{equation}
\end{small}
where  $d(\boldsymbol{p},\partial\mathcal{O}_i)$ is the minimum distance from $\boldsymbol{p}$ to the boundary of $\mathcal{O}_i$, noted as $\partial\mathcal{O}_i$. 
The closest point to $\boldsymbol{p}$ on $\partial\mathcal{O}_i$ is denoted as $\boldsymbol{o}_i^*$,  and the normal vector along the SDF direction can be calculated as follows:
\begin{small}
\begin{equation} 
     \boldsymbol{n} =sgn(sd(\boldsymbol{p},\mathcal{O}_i))\cdot (\boldsymbol{p}-\boldsymbol{o}_i^*)/\|\boldsymbol{p}-\boldsymbol{o}_i^*\|,\label{equ:cal normal verctor}
     \end{equation}
\end{small}
where $sgn(\cdot)$ is the sign function. 
The SDF and its corresponding normal vector $\boldsymbol{n}$ can be efficiently calculated using Gilbert–Johnson–Keerthi (GJK) algorithm \cite{gilbert1988fast} and Expanding Polytope Algorithm (EPA) \cite{van2001proximity}.

\subsection{Value-at-Risk and Conditional Value-at-Risk}
The value-at-risk (VaR) represents the minimum possible value of risk that can be achieved given a risk tolerance level $\alpha\in(0,1]$. 
Specifically, 
the VaR of a random variable $\zeta$ under $\alpha$ is defined as 
\begin{small}
\begin{equation}
    VaR_\alpha(\zeta)= \min \left\{ z\vert Pr(\zeta \leqslant z)\geqslant 1-\alpha \right\},
\end{equation}
\end{small}
where $Pr(\cdot)$ denotes the probability. 
The CVaR of a random variable $\zeta$ with a continuous distribution is defined as
\begin{small}
\begin{equation}
   CVaR_\alpha(\zeta)= \min\limits_{z\in\mathbb{R} } \mathbb{E} [ z+(\zeta-z)^+/\alpha]=\mathbb{E}[\zeta \vert  \zeta \geq VaR_\alpha(\zeta)],\label{equ:cvar_min_Def}
\end{equation}
    \end{small}
where $(\cdot)^+ $ = $\max(\cdot,0)$, and $\mathbb{E}$ is the expectation operator.

Let $\phi(\cdot)$ and $\Phi(\cdot)$ denote the PDF and the cumulative distribution function (CDF) of a standard normal distribution, respectively. The CVaR of a Gaussian random variable $\zeta\sim\mathcal{N}(\mu,{\sigma}^2)$ has the following closed-form expression \cite{norton2021calculating}
\begin{small}
\begin{equation}
    CVaR_\alpha(\zeta)=\mu+\sigma\phi(\Phi^{-1}(1-\alpha))/\alpha.
    \label{equ: Gaussian CVaR cal}
\end{equation}
\end{small}

\subsection{Problem Formulation}
\label{sec: Problem Formulation}
Consider a workspace $\mathcal{W}\subset\mathbb{R}^2$ that contains a robot swarm consisting of $N_r$ 
robots and $N_o$ obstacles $\mathcal{O}_i\subset\mathcal{W}, \forall i\in\underline{N_o}$. 
Let $\mathcal{O}=\bigcup_{i=1}^{N_o}\mathcal{O}_i$ represents the union of all obstacles. 
Obstacles are static and known a prior.
For simplicity, $\mathcal{O}_i$ is assumed to be convex. 
For non-convex obstacles, the convex hull or convex decomposition can be applied to obtain convex obstacles.

The proposed swarm motion planning problem aims to safely transport robots from an initial area to a target region in obstacle-cluttered environments (\cref{fig:coverr}). 
We propose to address this task through a hierarchical strategy involving a macroscopic planning stage and a microscopic control stage. 

At the macroscopic level, the swarm state can be represented by a time-varying random variable $\boldsymbol{X}(k)$ with $k$ denoting the time step. The PDF of $\boldsymbol{X}(k)$ indicates the density distribution of robots across $\mathcal{W}$. 

Considering the universal approximation property of Gaussian Mixture Models (GMMs), we choose a GMM $\wp_k=\sum_{j=1}^{N_k}\omega_k^j g^j_k$ with $ \sum_{j=1}^{N_k}\omega_k^j =1$ from the GMM distribution space $\mathcal{G}(\mathcal{W})$ to represent the macroscopic state of the robot swarm, i.e., $\boldsymbol{X}(k)\sim\wp_k$.
Here $N_k$ denotes the number of GCs in the GMM and $g^j_k$ is the $jth$ Gaussian distribution with mean $\boldsymbol{\mu}^j_k$, covariance matrix $\boldsymbol{\Sigma}^j_k$ and weight $\omega^j_k\geq 0$. 

The macroscopic planning stage is formulated as a swarm PDF OT problem (P0) from an initial distribution $\wp_0\in\mathcal{G}(\mathcal{W})$ to a target distribution $\wp_{targ}\in\mathcal{G}(\mathcal{W})$ while minimizing the total transport cost and satisfying a set of collision avoidance constraints, as follows:
\vspace{-0.3cm}
\begin{small}
\begin{subequations}\label{equ:OT}
\begin{align}\label{equ: OC formulation}
\min\limits_{\wp_{1},\wp_{2},\dots,\wp_{T_f}} J& \triangleq \sum_{k=0}^{T_f-1} \lambda_k W_2(\wp_k,\wp_{k+1})\\
s.t.\quad & CVaR_\alpha(-sd(\boldsymbol{X}(p),\mathcal{O}_i))<\epsilon, \label{con: collision avoidance}\\& f_l(\wp_p)<\varepsilon,\label{con: max PDF}\\&\forall p\in \underline{T_f},\forall i\in \underline{N_o},
\end{align}
\end{subequations}  
\end{small}
where the optimization objective $J$ is the sum of the Wasserstein distance between consecutive PDFs from the initial time step to the terminal time step $T_f$ when $\wp_{T_f}=\wp_{targ}$, with corresponding weight coefficients $\lambda_k>0$.
The constraint \cref{con: collision avoidance} limits the expectation of SDF between the swarm PDF and obstacles. 
Taking the opposite sign of SDF is performed to accommodate the definition of CVaR. Details regarding $sd(\cdot,\cdot)$ will be provided in \cref{sec:Linearization of SDF}.
The constraint \cref{con: max PDF} with a linear function $f_l(\cdot)$ and a probability bound $\varepsilon\in(0,1)$ ensures that the robots do not come too close to each other. 
A more detailed elaboration of (P0) will be presented in \cref{sec.Macroscopic Formulation}.

From the microscopic aspect, the position of every single robot $\boldsymbol{x}_i(k)$ can be treated as a sample from the swarm GMM, with a motion model formulated as:
\begin{small}
\begin{equation} 
 \begin{aligned}
     & \boldsymbol{x}_i(k+1) = f(\boldsymbol{x}_i(k), \boldsymbol{u}_i(k)),\\ & k=0,1,\cdots,T_f-1, 
\forall i\in\underline{N_r},
 \end{aligned}
 \end{equation}
\end{small}
where $f(\cdot,\cdot)$ denotes the robot motion function, $\boldsymbol{x}_i(k)$ and $\boldsymbol{u}_i(k)$ denote the state and control input of every single robot at time step $k$. 
At the microscopic control stage, each robot computes its own control inputs by the APF method to keep track of the swarm PDF obtained from the macroscopic planning level and avoid collisions with static obstacles and other robots. 
The microscopic control falls beyond the scope of this work, and interested readers are encouraged to refer to  \cite{zhu2019scalable} for details.

\begin{figure}[t]
\centering
\includegraphics[width=1\linewidth]{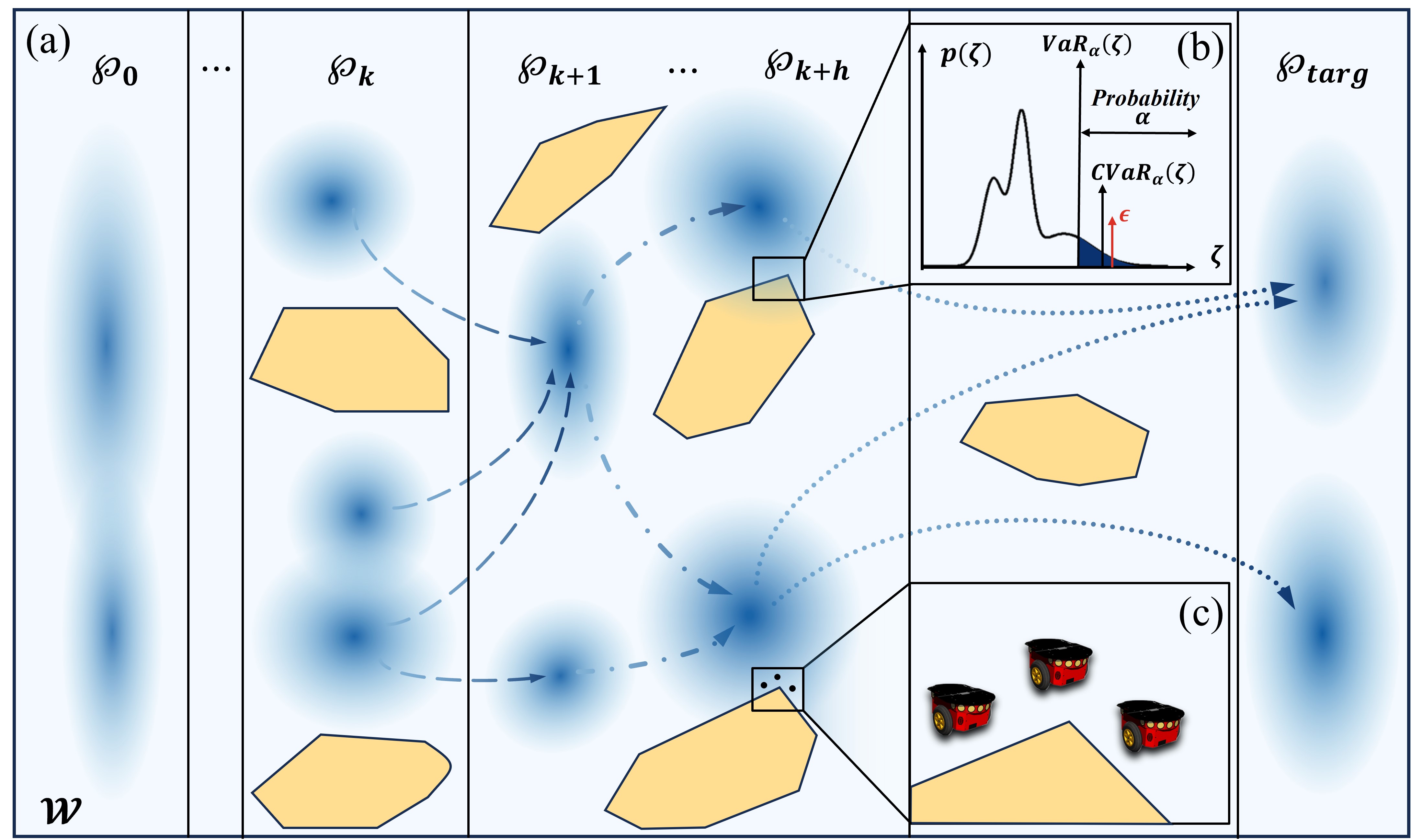}
\caption{Illustration of ROVER transporting the robot swarm from an initial area in the left column to the target area in the right column in a cluttered workspace $\mathcal{W}\subset\mathbb{R}^2$. 
Static obstacles are denoted by pale yellow polygons. 
Columns separated by verticle black lines correspond to different time steps. 
(a) \textbf{Macroscopic planning in ROVER.} 
Blue clouds represent the GMM distributions $\wp$ of the swarm state. 
Blue dashed lines illustrate the GMM transformation from $\wp_k$ to $\wp_{k+1}$, while the dash-dotted lines represent $(h-1)$ subsequent transformations from $\wp_{k+1}$ to $\wp_{k+h}$. 
The dotted lines represent the transformation from $\wp_{k+h}$ to $\wp_{targ}$. 
(b) \textbf{Collision avoidance between GMM and obstacles using CVaR.} 
The random variable $\zeta$ represents the distribution of distance between a GMM and obstacles.
The shaded area denotes the $\alpha \%$ of the area under $p(\zeta)$. 
The collision avoidance under a risk acceptance level $\alpha\in(0,1]$ is to constrain the $CVaR_\alpha(\zeta)$, the expected value of $\zeta$ under the shaded area, below a user-defined threshold $\epsilon$. 
(c) \textbf{Microscopic control in ROVER.}
Robots track the GMM trajectories given from the macroscopic level while avoiding collision.}
\label{fig:method}
\end{figure}

\section{Methodology}
This section provides a comprehensive outline of the methodology, including the detailed formulation of macroscopic planning (\cref{sec.Macroscopic Formulation}), the SDF distribution between swarm GMM and obstacles (\cref{sec:Linearization of SDF}), the formulation of GMM-CVaR collision avoidance constraints (\cref{sec:GMM-CVaR}), followed by an online implementation with SLP framework (\cref{sec:SLP}). 
A schematic overview is provided in \cref{fig:method}. 

\subsection{Macroscopic Planning Formulation}\label{sec.Macroscopic Formulation}
\begin{figure}[!t]
\centering
\includegraphics[width=8.5cm]{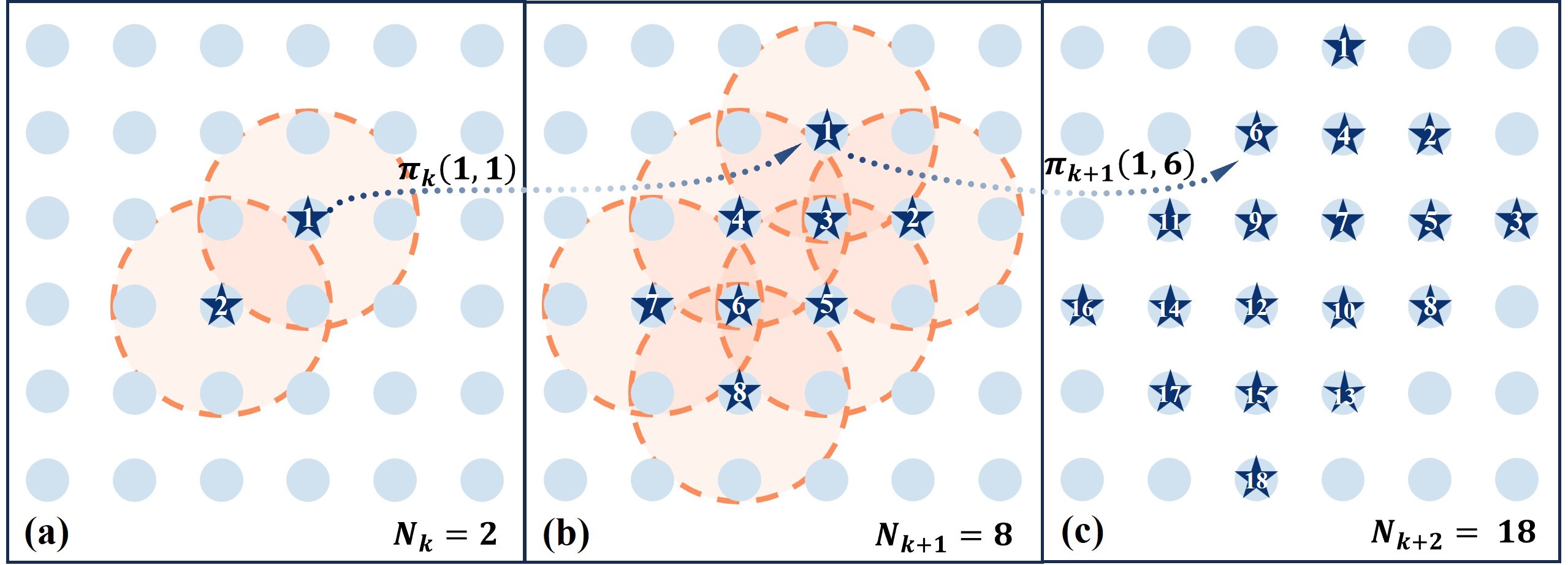}
\caption{An example of determining $\mathcal{GC}_k$ and $N_k$, $\forall k\in \underline{T_f}$, based on \cref{ass:GC-set} and \cref{ass:N_GC}. 
Figures (a), (b), and (c) represent GMMs at time steps $k$, $k+1$, and $k+2$, respectively. Light blue circles depict GCs in $\mathcal{C}$ and pentagrams with numbers represent the GCs at specific time step and their corresponding indices. 
The dotted lines give an example of weight transported between consecutive time steps within a transport range denoted by light orange circles. 
}
\label{fig:collocation_point}
\end{figure}

The proposed (P0) is an infinite-dimensional optimization problem, which is computationally prohibitive.
Thus, we proposed two approximations to simplify the problem. 

In the first approximation, we uniformly discretize the planning space and establish a set $\mathcal{C}$ of GCs whose mean positions are constrained on the discrete points and the covariance matrices are uniform. 
We then propose the following assumptions:

\begin{assumption} \label{ass:GC-set}
All GCs of the GMM in the planning stage can only be chosen from $\mathcal{C}$, i.e., $g^j_k \in \mathcal{C}$, $\forall j \in\underline{N_k},\forall k\in\underline{T_f}$.
\end{assumption}
\begin{assumption}\label{ass:N_GC}
Between consecutive planning steps, GCs can be only transported to the GCs within a predefined transport radius measured by the Wasserstein Metric.
\end{assumption}

Leveraging \cref{ass:GC-set} and \cref{ass:N_GC}, a set of all GCs at time step $k$ denoted as $\mathcal{GC}_k$ and $N_k=|\mathcal{GC}_k|$ can be predetermined, $\forall k\in \underline{T_{f}}$. 
Only weights of the GMMs remain to be optimized. 
A schematic explanation of these assumptions can be found in \cref{fig:collocation_point}. 

In the second approximation, we propose to reduce the dimension of (P0) utilizing an FTMPC framework. 
The transport cost is composed of stage cost and terminal cost. 
The stage cost represents the transport cost within the FTMPC planning horizon, and terminal cost is defined as the cost from the last step in the planning horizon to $T_f$. 
To calculate the terminal cost, we make the following assumption:
\begin{assumption}\label{ass:cost_split}
The GCs of the GMM from the last step in the planning horizon to $T_f-1$ do not split or merge.
Transformation into the target distribution is conducted in the terminal time step $T_f$. 
\end{assumption}

With the help of these two approximations, (P0) is reformulated into an FTMPC problem (P) as follows:

  \begin{subequations}
  \footnotesize
    \begin{align}
    \label{equ:cost_function}
   \min\limits_{\boldsymbol{\pi}} \ J
     &\triangleq \sum_{p=1}^{h} \sum_{i=1}^{N_{k+p-1}} \sum_{j=1}^{N_{k+p}} \lambda_{k+p-1} W_2^2(g_{k+p-1}^i, g_{k+p}^j) \pi_{k+p-1}(i,j)
   \notag\\ 
&+\sum_{m=1}^{N_{k+h}} \sum_{n=1}^{N_{targ}} \lambda_{k+h} \mathcal{Q}_{k+h}(m,n) \pi_{k+h}(m,n)\\
    s.t. 
\quad&\omega_{k+p-1}^i=\sum_{j=1}^{N_{k+p}}\pi_{k+p-1}(i,j),\  \forall i\in\underline{N_{k+p-1}} \label{equ:marginal PMF1} 
\\
&\quad\omega_{k+p}^j=\sum_{i=1}^{N_{k+p-1}}\pi_{k+p-1}(i,j), \  \forall j\in\underline{N_{k+p}} 
\label{equ:marginal PMF2}\\
&\quad\omega_{k+h}^m=\sum_{n=1}^{N_{targ}}\pi_{k+h}(m,n), \  \forall m\in\underline{N_{k+h}} 
\label{equ:marginal PMF3}\\ 
&\quad\omega_{targ}^n=\sum_{m=1}^{N_{k+h}}\pi_{k+h}(m,n), \forall n\in\underline{N_{targ}}
\label{equ:marginal PMF4}\\ 
&\quad\sum_{m=1}^{N_{k+p}}\omega_{k+p}^m=\sum_{n=1}^{N_{targ}}\omega_{targ}^n=1\\
& \quad \boldsymbol{\omega}_{k+p}\geq\boldsymbol{0}, \boldsymbol{\omega}_{targ}\geq\boldsymbol{0}\\
&\quad \sum_{i=1}^{N_{k+p-1}}\sum_{j=1}^{N_{k+p}}f_l(g_{k+p}^j)\pi_{k+p-1}(i,j)<\varepsilon
\label{con:maxPDF1}\\
&\quad CVaR_\alpha(-sd(\boldsymbol{X}(k+p),\mathcal{O}_\iota))<\epsilon, \label{con:cvar-gmm}\\
&\quad \forall p\in\underline{h},\ \forall\iota\in\underline{N_o}\label{con:counter1}
\end{align}
\end{subequations}
where $h$ represents the planning horizon. 
The transport policy $\boldsymbol{\pi}=[\pi_k(\cdot,\cdot),\pi_{k+1}(\cdot,\cdot),\cdots,\pi_{k+h}(\cdot,\cdot)]$ is a stack of joint Probability Mass Functions (PMFs) whose marginal PMFs are denoted by \cref{equ:marginal PMF1,equ:marginal PMF2,equ:marginal PMF3,equ:marginal PMF4}. 
As an example, $\pi_k(i,j)$ represents the weight transported from $g^i_k$ to $g^j_{k+1}$. 
The optimal swarm PDF at the next time step is obtained by
\begin{small}
\begin{equation}
\wp_{k+1}^*=\sum\limits_{j=1}^{N_{k+1}}\sum\limits_{i=1}^{N_k} \pi_k^*(i,j)g_{k+1}^j.\label{equ:next_pdf}
\end{equation}    
\end{small}

The former half of \cref{equ:cost_function} represents the stage cost with weights $\lambda_{k:k+h-1}$, while the latter half represents the terminal cost with a weight $\lambda_{k+h}$. 
Under \cref{ass:GC-set} and \cref{ass:N_GC}, 
the squared terms of $W_2(\cdot,\cdot)$ in the stage cost can be calculated offline. 
 For the terminal cost, we adopt the transport policy as in \cref{ass:cost_split} and establish a directed graph $\mathbf{G}$ whose vertices are GCs from $\mathcal{C}$ and weights are determined by the transport distance and the collision risk between two vertices.
The cost term $\mathcal{Q}_{k+h}(m,n)$ in the terminal cost can be computed by applying a shortest-path-planning algorithm on $\mathbf{G}$ in advance. 
Therefore, the cost function \cref{equ:cost_function} is linear w.r.t. $\boldsymbol{\pi}$.

Similarly, the constraint \cref{con:maxPDF1} is linear w.r.t. $\boldsymbol{\pi}$. 
But the formulation of constraint \cref{con:cvar-gmm} remains unclear, which will be elaborated in \cref{sec:Linearization of SDF} and \cref{sec:GMM-CVaR}. 

\subsection{SDF under GMM Representation}\label{sec:Linearization of SDF}
The collision avoidance constraint \cref{con:cvar-gmm} involves SDF between the swarm GMM and obstacles. 
We consider a GMM random variable $\boldsymbol{X}\sim\sum_{j=1}^N \omega^j\mathcal{N}(\boldsymbol{\mu}^j,\boldsymbol{\Sigma}^j)$ and derive the SDF distribution between $\boldsymbol{X}$ and an obstacle noted as $sd(\boldsymbol{X},\mathcal{O}_i)$.

For simplicity, we first consider the $jth$ GC of the GMM denoted as $\boldsymbol{X}^j\sim\mathcal{N}(\boldsymbol{\mu}^j,\boldsymbol{\Sigma}^j)$ and its corresponding $sd(\boldsymbol{X}^j,\mathcal{O}_i)$.
Utilizing the EPA or GJK algorithm and \cref{equ:cal normal verctor}, the deterministic SDF between the mean position of $\boldsymbol{X}^j$ and an obstacle $\mathcal{O}_i$, denoted by $sd(\boldsymbol{\mu}^j,\mathcal{O}_i)$, and the corresponding normal vector $\boldsymbol{n}^j$, can be calculated.

The $sd(\boldsymbol{X}^j,\mathcal{O}_i)$ can be subsequently approximated by the first-order Taylor expansion as follows,
\begin{small}
 \begin{equation}
sd(\boldsymbol{X}^j,\mathcal{O}_i)\approx sd(\boldsymbol{\mu}^j,\mathcal{O}_i)+\nabla{sd(\boldsymbol{X}^j,\mathcal{O}_i)} \vert _{\boldsymbol{X}^j=\boldsymbol{\mu}^j}(\boldsymbol{X}^j-\boldsymbol{\mu}^j),\label{equ:linearized SDF}
 \end{equation}
 \end{small}
where $\nabla{sd(\boldsymbol{X}^j,\mathcal{O}_i)}\vert _{\boldsymbol{X}^j=\boldsymbol{\mu}^j}=(\boldsymbol{n}^{j})^T$.
The reasonableness of this approximation stems from the assumption that when calculating $sd(\boldsymbol{X}^j,\mathcal{O}_i)$, the closest point $\boldsymbol{o}_i^*$ and the normal vector $\boldsymbol{n}^j$ remain unchanged for any sample of  $\boldsymbol{X}^j\sim\mathcal{N}(\boldsymbol{\mu}^j,\boldsymbol{\Sigma}^j)$ \cite{yang2023risk}. 
The approximation error is acceptable when samples of $\boldsymbol{X}^j$ lie in the neighbourhood of $\boldsymbol{\mu}^j$. 
Notice that $sd(\boldsymbol{X}^j,\mathcal{O}_i)$ is a Gaussian random variable as a result of the linear transformation of $\boldsymbol{X}^j$, i.e.,
\begin{small}
\begin{equation}
sd(\boldsymbol{X}^j,\mathcal{O}_i)\sim\mathcal{N}(sd(\boldsymbol{\mu}^j,\mathcal{O}_i),(\boldsymbol{n}^j)^T\boldsymbol{\Sigma}^j\boldsymbol{n}^j).\label{equ:gaussian SDF}
\end{equation}    
\end{small}
We further derive the distribution of $sd(\boldsymbol{X},\mathcal{O}_i)$ based on \cref{equ:gaussian SDF} and the following proposition:
\begin{proposition}\label{pro:sdf_distribution}
 The distribution $\sum\limits_{j=1}^N \omega^j Pr(sd(\boldsymbol{X}^j,\mathcal{O}_i))$ and distribution $Pr(sd(\boldsymbol{X},\mathcal{O}_i))$ are equal.

The proof can be found in \cref{appen:SDF of GMM}.
\end{proposition}

\Cref{pro:sdf_distribution} yields that $sd(\boldsymbol{X},\mathcal{O}_i)$ follows a GMM distribution as follows:
\begin{small}
\begin{equation}
sd(\boldsymbol{X},\mathcal{O}_i)\sim\sum_{j=1}^N \omega^j\mathcal{N}(sd(\boldsymbol{\mu}^j,\mathcal{O}_i),(\boldsymbol{n}^j)^T\boldsymbol{\Sigma}^j\boldsymbol{n}^j).
\end{equation}
\end{small}

Therefore, the $sd(\cdot,\cdot)$ in \cref{con:cvar-gmm} can be calculated by $
sd(\boldsymbol{X},\mathcal{O}_i)= \sum_{j=1}^N \omega^j sd(\boldsymbol{X}^j,\mathcal{O}_i)
$ utilizing \cref{equ:linearized SDF}. 

\subsection{Analytical Expression of GMM-CVaR}\label{sec:GMM-CVaR}
The collision avoidance constraints \cref{con:cvar-gmm} involve GMM-CVaR, for which there is currently no analytical expression available. 
Therefore, we first propose a proposition concerning GMM-CVaR.

\begin{proposition}
\label{pro:cvar-gmm expression}
    The CVaR of a GMM random variable $Y$ at risk acceptance level $\alpha$ can be represented by the summation of CVaRs of the corresponding Gaussian component $Y^j$ of $Y$ at the risk acceptance level $\alpha_j$:
    \begin{small}
    \begin{equation}
        CVaR_{\alpha}(Y) =\frac{1}{\alpha}\sum\limits_{j=1}^N \omega^j \alpha_j CVaR_{\alpha_j}(Y^j),\label{equ:CVaR_GMM}
    \end{equation}
    \end{small}
    where $\omega^j$ denotes the weight of $Y^j$, and $\alpha_j$ is the tail probability of the $Y^j$ distribution at the $VaR_\alpha(Y)$ quantile, i.e., 
$\alpha_j=\int_{VaR_\alpha(Y)}^{+\infty} p(Y^j) dy$. 
Besides, there exists relationship between $\alpha$ and $\alpha_j$ denoted as $\alpha=\sum_{j=1}^N \omega_j \alpha_j$.

The proof is given in \cref{appen:CVaR-GMM expression}.
\end{proposition}

Thus, the constraint \cref{con:cvar-gmm} can be explicitly expressed by substituting $Y$ in \cref{pro:cvar-gmm expression} with $Y(k+p)=-sd(\boldsymbol{X}(k+p),\mathcal{O}_\iota)$ and $Y^j$ in \cref{pro:cvar-gmm expression} with $Y^j(k+p)=-sd(\boldsymbol{X}^j(k+p),\mathcal{O}_\iota)$, $\forall \iota\in\underline{N_o}$:
\vspace{0cm}
\begin{subequations}
\footnotesize
\begin{align}
     &CVaR_{\alpha}(Y(k+p)) =\frac{1}{\alpha}\sum\limits_{j=1}^{N_{k+p} }\omega_{k+p}^j \alpha_j CVaR_{\alpha_j}(Y^j(k+p))<\epsilon,
     \notag\\ 
     &\forall p\in\underline{h}.
     \label{equ:gmm_cvarandGaussianCVaR}
\end{align}
\end{subequations}

Although $CVaR_{\alpha_j}(Y^j(k+p))$ can be explicitly expressed by \cref{equ: Gaussian CVaR cal}, $\alpha_j$ relating with $VaR_\alpha(Y(k+p))$ is an implicit function of the weights of GCs, which renders \cref{con:cvar-gmm} a nonlinear constraint.
In the next section, we propose a computationally efficient algorithm customized to (P) by leveraging the structure of \cref{equ:gmm_cvarandGaussianCVaR}.

\subsection{Online implementation via Sequential Linear Programming (SLP)}\label{sec:SLP}
Since the constraint \cref{con:cvar-gmm} turns out to be the only nonlinear term in (P), we propose to linearize the constraint \cref{con:cvar-gmm} to achieve online implementation of ROVER. 
Specifically, we adopt an SLP framework \cite{griffith1961nonlinear} and seek to solve the optimization problem (P) by solving a series of linear programming (LP) subproblems (P$^v$). 
 For each subproblem (P$^v$), we linearize the $CVaR_\alpha(Y(k+p))$ constraint using the first-order Taylor expansion at a feasible solution of (P$^v$) $\boldsymbol{\pi}^v$ as
 
\vspace{-0.35cm}
\begin{small}
\begin{subequations}
\begin{align}\label{equ:linearized CVaR GMM}
&CVaR_\alpha(Y(k+p))=CVaR_\alpha(Y_v(k+p))\notag\\
     &+\nabla ^T_{\boldsymbol{\pi}} CVaR_\alpha(Y(k+p))|_{\boldsymbol{\pi}=\boldsymbol{\pi}^v}(\boldsymbol{\pi}-\boldsymbol{\pi}^v),
  \end{align}
\end{subequations}
\end{small}
where $Y_v(k+p)$ denotes the SDF random variable $Y(k+p)$ under a given $\boldsymbol{\pi}^v$, and its corresponding $CVaR_\alpha(Y_v(k+p))$ can be calculated with \cref{equ:gmm_cvarandGaussianCVaR}.
It should be noted that the $\boldsymbol{\pi}$ used in this section represents the flattened vector form of the matrix form $\boldsymbol{\pi}$ in \cref{sec.Macroscopic Formulation}.

The subproblem (P$^v$) at each iteration is as follows:
\begin{small}
  \begin{subequations}
    \begin{align}
  & \min\limits_{\boldsymbol{\pi}} \quad J
    \label{equ:obj_aug}
   \\
   s.t. 
 &\quad\eqref{equ:marginal PMF1}-\eqref{con:maxPDF1},\eqref{con:counter1},
\\
& \quad\eqref{equ:linearized CVaR GMM}<\epsilon,\label{con:linearized CVAR gmm} 
\\
& \quad -\boldsymbol{s}^v<\boldsymbol{\pi}-\boldsymbol{\pi}^v<\boldsymbol{s}^v,\label{con:step_size SLP} 
\end{align}
\end{subequations}
\end{small}
where \cref{con:step_size SLP} is an element-wise inequality constraint and each dimension of $\boldsymbol{s}^v$ is a step bound parameter $s^v>0$. This constraint ensures that the feasible domain of (P$^v$) remains in the neighbourhood of $\boldsymbol{\pi}^v$, where linearization occurs.

The detailed execution of the proposed $\textsc{OptiGMM}$ within an SLP framework is presented in \cref{alg:OptiGMM}. 
Every iteration involves first the acquisition of GMM $Y(k+p)$ with the knwoledge of $\boldsymbol{\omega}_{k+p}$ (\cref{code:cal weight from pi}) and $\mathcal{GC}_{k+p}$. \Cref{code:cvar_Linearization} encompasses the linearization of $CVaR_\alpha(Y(k+p))$. \Cref{code:solve Pv} to \cref{code:step size procedure} incorporates solving (P$^v$) to find the optimal iterative solution $\boldsymbol{\pi}^{v*}$ and update $\boldsymbol{\pi}^v$ and $\boldsymbol{s}^v$. 
This algorithm is sequentially executed until $\boldsymbol{\pi}^v$ meets the convergence condition.
\begin{algorithm}
\fontsize{10pt}{10pt}\selectfont
  \caption{$\textsc{OptiGMM}$}
  \label{alg:OptiGMM}
  \begin{algorithmic}[1]
  \Statex \textbf{Input:} $\boldsymbol{\pi}^0, \mathcal{GC}_{k+1:h}, \mathcal{O}, \alpha\in (0,1], \eta>0, s^0>0$
\State Initialization: $v \leftarrow 0$
\While{$\lVert\boldsymbol{\pi}^{v}-\boldsymbol{\pi}^{v-1}\rVert>\eta$ or $v==0$}
 \For{$i = 1$ to $N_o$}
  \For{$p = 1$ to $h$}
   \State compute $\boldsymbol{\omega}_{k+p}$ from $\boldsymbol{\pi}^v$ based on \cref{equ:marginal PMF2} \label{code:cal weight from pi}
     \State linearize $CVaR_\alpha(Y(k+p))$ based on \cref{equ:linearized CVaR GMM} \label{code:cvar_Linearization}
     \EndFor
     \EndFor
    \State solve (P$^v$) to find $\boldsymbol{\pi}^{v*}$ \label{code:solve Pv}
    \State update the step bound $\boldsymbol{s}^{v+1}$ 
    \State $\boldsymbol{\pi}^{v+1} \leftarrow \boldsymbol{\pi}^{v}$ \label{code:step size procedure}
    \State $v \leftarrow v+1 $
   \EndWhile 
   \State compute $\wp_{k+1}^*$ from $\boldsymbol{\pi}^{v*}$ based on \cref{equ:next_pdf}
  \Statex \textbf{Output:} $\wp_{k+1}^*$ 
  \end{algorithmic}
\end{algorithm}

\section{Simulation and Results}

In this section, we evaluate the performance of ROVER via simulations. 
The workspace $\mathcal{W}$ is a $[0,200]\times[0,160]m^2$ area with static obstacles. 
The initial distribution of the swarm consists of four GCs with a shared covariance matrix $\boldsymbol{\Sigma}=100\boldsymbol{I}_2$, where $\boldsymbol{I}_2$ denotes a $2\times 2$ identity matrix. 
Other parameters are $\boldsymbol{\mu}_1=[25,35]$, 
$\omega_1=0.25$, $\boldsymbol{\mu}_2=[25,55]$,  $\omega_2=0.375$, $\boldsymbol{\mu}_3=[25,115]$, $\omega_3=0.1875$, $\boldsymbol{\mu}_4=[25,135]$, $\omega_4=0.1875$. 
The target distribution is composed of three GCs with parameters $\boldsymbol{\mu}_1=[175,120]$,  $\omega_1=0.25$, $\boldsymbol{\mu}_2=[175,60]$, $\omega_2=0.375$, $\boldsymbol{\mu}_3=[175,40]$,  $\omega_3=0.375$ and an identical covariance matrix $\boldsymbol{\Sigma}=100\boldsymbol{I}_2$. 
The GC set $\mathcal{C}$ is predefined by taking the mean of each GC on fixed grids, where the X coordinates range from 5 to 195 and the Y coordinates range from 5 to 155, with a discretization interval of 10. 
Consequently, the set $\mathcal{C}$ comprises a total of $20\times16=320$ GCs, each with a same covariance matrix $\boldsymbol{\Sigma}_c=50\boldsymbol{I}_2$. 
We set the discretization time interval $\Delta t$ to $0.1s$.
Robots are assumed to adopt omnidirectional models with a maximal speed of $1.5m/s$ and the radius of each robot is set to $0.12m$. 
To specify the optimization problem, $\alpha$ and $\epsilon$ are set to 0.05 and -0.2 respectively. 
In the objective function, we set $h$ to 2, $\lambda_{k+h}$ to 3, and $\lambda_{k+p-1}$ to 1, $\forall p\in\underline{h}$.
In \cref{alg:OptiGMM}, 
we adopt a fixed step bound as $s^v = 0.1, \forall v\in\mathbb{N}_+$. 
The $\eta$ in the convergence condition is $10^{-5}$.
All simulations are run on a desktop (13th Intel(R) i7 CPU@2.10GHz) and LP is solved using the ``mosek'' solver with interior-point algorithm in MATLAB.

We propose five metrics to quantify the performance of a motion planner: 
\begin{itemize}
     \item Total runtime $t$ excluding the offline data preparation phase. $t$ is set to be infinite when the swarm transport task fails to be completed within $T_f^{max}=3000$. 
    \item Average runtime per time step $t_f=t/T_f$.
    \item Average trajectory length $\Bar{D}$ until reaching the goal. 
    \item Minimum inter-robot distance $min(d_{ij})$ and minimum robot-obstacle distance $min(d_{io})$ maintained by all robots over the whole trajectory.

\end{itemize}
\subsection{Comparison with Benchmark Approaches}

We compared the performance of our method ROVER with several state-of-art approaches in a more complex scenario characterized by a host of fragmented polygonal obstacles(\cref{fig:traj}). 
Approaches highly pertinent to our research include multi-robot Formation Control (FC) \cite{alonso2017multi}, Predictive Control (PC) \cite{soria2021predictive,soria2021distributed}, and Optimal Tube swarm planning and control (OT) \cite{mao2022optimal}. 
However, OT involving a single virtual tube planning cannot accomplish our swarm transport task, where our swarm is partitioned into distinct subgroups for various goal areas concerning variation in swarm density.
Therefore, the following section presents a comparative analysis of ROVER, FC and PC. 
Notice that FC is a sampling-based method, the numerical results of FC are mean values obtained from five random simulations. 
Furthermore, PC utilizes soft constraints for collision avoidance, allowing robots to continue moving towards target positions even after collisions occur. 
We approximate polygonal obstacles using circular obstacles in the simulation of PC to accommodate the collision avoidance mechanism of PC.

 \subsubsection{Flexibility Analysis of Different Approaches} \label{subsubsec:Adaptability to Different Environments}
\begin{figure*}[!h]
\centering
\includegraphics[width=17.5cm]{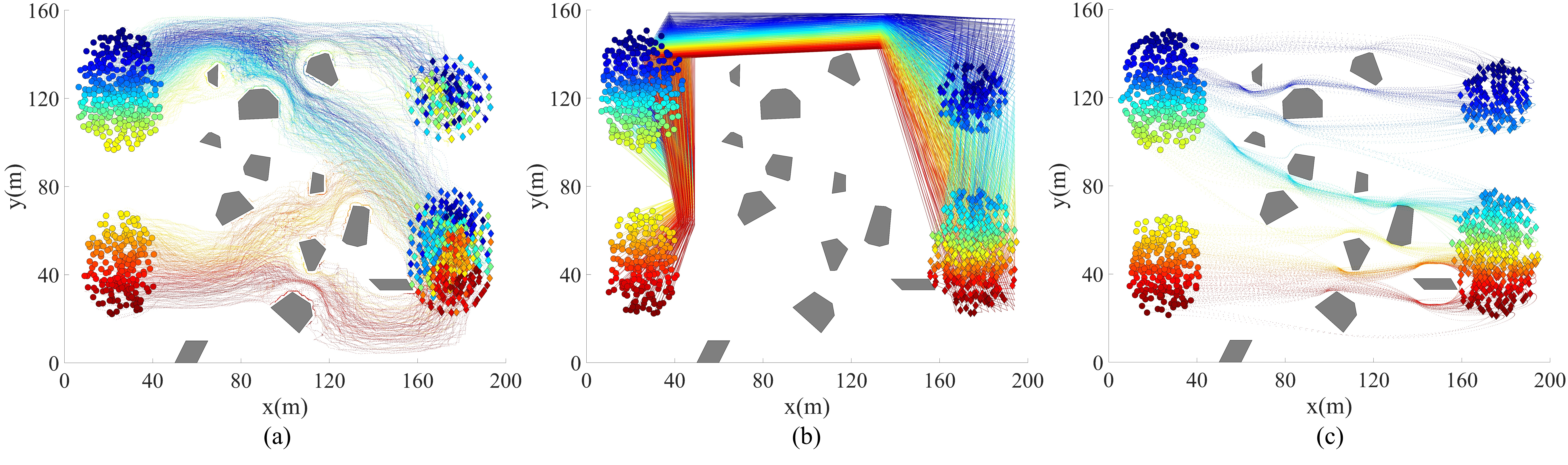}
\caption{Figure (a)-(c) shows the trajectories of 500 robots generated by ROVER, FC, and PC, respectively. The initial positions of robots are demonstrated by circles, while their corresponding final positions are highlighted by diamonds. 
The grey polygons denote static obstacles.}
\label{fig:traj}
\end{figure*}

A comparison with benchmark approaches with a fixed swarm size $N_r=500$ is presented in this subsection.
The trajectories of robots are illustrated in \cref{fig:traj}.

\begin{table*}[!t]
\centering
\renewcommand{\arraystretch}{1} 
\caption{ }
\label{tab:scalability}
\begin{tabularx}{\textwidth}{@{} 
  >{\centering\arraybackslash}m{1.4cm} 
  >{\centering\arraybackslash}m{3.5cm} 
  >{\centering\arraybackslash}m{2.5cm}
  >{\centering\arraybackslash}m{3cm} 
  >{\centering\arraybackslash}m{2.5cm} 
  >{\centering\arraybackslash}m{2.8cm} 
 @{}}
\toprule
   $N_r$ &$t(min)$ &$t_f(s)$& $\Bar{D}(m)$ &$min(d_{ij})(m)$ & $min(d_{io})(m)$\\
\midrule
25
& 1.85 (0.66) / 2.35 / 1.01
& 0.21 / 0.05 / 0.33
& 223.00 / 345.40 / 165.83
& 1.03 / 0 / 0.29
& 0.38 / 16.63 / 0.09
 \\
50
& 2.10 (0.67) / 3.55 / 1.71
& 0.22 / 0.09 / 0.55
& 210.87 / 305.19 / 158.35
& 0.84 / 0.44 /	0.06
& 0.37 / 11.35 / 0.05
 \\
250
& 3.46 (0.67) / 24.12 / 22.10
& 0.36 / 0.58 / 6.20
& 202.81 / 323.31 / 160.60
& 0.38 / 0 / -0.2
& 0.23 / 8.95 / -2.71 \\
500
& 6.05 (0.67) / 58.58 / 120.78
& 0.64 / 1.28 / 31.51
& 202.26 / 364.16 / 159.036
& 0.34 / 0 / -0.21
& 0.22 / 6.68 / -3.08
 \\
1000
& 10.89 (0.66) / NaN / 564.05
& 1.09 / NaN / 120.01
& 204.37 / NaN / 159.42
& 0.32 / NaN / -0.22
& 0.17 / NaN / -4.10
 \\
\bottomrule
\end{tabularx}
\end{table*}

\cref{fig:traj} showcases the flexibility of different methods. 
Through planning and control at the individual level, the microscopic approach PC exhibits the utmost flexibility.  
As hierarchical approaches, ROVER demonstrates greater flexibility compared to FC. ROVER enables the swarm to split and merge when encountering obstacles, while FC constrains the swarm to travel as a group, compromising its flexibility.
The flexibility of ROVER benefits from the GMM representation of a swarm at the macroscopic level. 
The adaptive change of GCs and their corresponding weights in the GMMs facilitates the swarm's splitting and merging behaviour, which results in likewise flexibility as provided by the microscopic method PC.

\subsubsection{Scalability in Various Swarm Size}
We conduct a comparison with varying swarm sizes and the results in \cref{tab:scalability} are presented in the manner of ROVER / FC / PC, respectively. 
The time results in the parentheses denote the aggregated macroscopic planning duration of ROVER. 
FC fails to satisfy the inter-robot distance constraints when $N_r=1000$, leading to the NaN results. 

In terms of computational efficiency ($t$ and $t_f$), 
PC outperforms the other two approaches at smaller swarm scales ($N_r=25,50$). 
However, as the swarm size increases, the advantages of hierarchical methods (ROVER and FC) become increasingly significant, while the computational cost of PC rises sharply. 
Additionally, ROVER exhibits superior performance than FC across various swarm sizes. 
It is worth noticing that ROVER only requires a runtime of around $10min$ when $N_r=1000$ and can reach the $1Hz$ control frequency, enabling online implementations. 
The remarkable scalability of ROVER is attributed to the complete decoupling of the macroscopic planning level from the microscopic control phase. 
Robots in our work can be viewed as samples from swarm PDFs. 
Thus, increasing the swarm size does not affect macroscopic PDF planning, which corresponds to the stable planning duration within the parentheses.

Regarding the average trajectory length ($\Bar{D}$), ROVER exhibits a decreasing trend as the swarm scale increases. 
This can be explained by the fact that the PDF paths provided by ROVER's macroscopic planning are similar across various swarm sizes, and the growth in robot density forces peripheral robots in the swarm to choose trajectories closer to obstacles but with shorter distances. 
Conversely, the trajectory length of FC is the utmost and does not follow a discernible pattern since FC only samples a feasible formation sequence without guaranteeing optimality in travel distance. 
PC maintains a relatively stable and shortest path across all swarm sizes because PC optimizes every robot's path individually with fixed start and goal positions.

As for collision avoidance, ROVER exhibits a decrease in the distance to obstacles and among robots as the robot density surges, while still guaranteeing collision-free trajectories. 
FC demonstrates the most conservative obstacle avoidance performance but reduces the inter-robot distance ($min(d_{ij}=0)$ in $N_r=250,500$). 
PC encounters collisions (negative $min(d_{ij})$ and $min(d_{io})$) except for small swarm scales ($N_r=25,50$) due to soft collision avoidance constraints.

\section{Conclusion}
In this work, we propose ROVER for risk-aware motion planning of large-scale robotic swarms.
The GMM representation of the swarm brings great flexibility to ROVER, allowing the robot swarm to split and merge to travel around obstacles. 
The SLP framework allows ROVER to conduct online motion planning for large-scale swarm robotic systems. 
The incorporation of CVaR enables effective risk management between the swarm and obstacles, ensuring safe navigation through cluttered environments.
 Future work will extend ROVER to unknown environments and evaluate the approach using real-world experiments.


\bibliographystyle{ieeetr}
\bibliography{references}

\section*{APPENDIX}
\subsection{Proof of \cref{pro:sdf_distribution}}\label{appen:SDF of GMM}
\begin{proof}
This section is dedicated to building the relationship between the distributions $\sum\limits_{j=1}^N \omega_j Pr(-sd(\boldsymbol{X}^j,\mathcal{O}_i))$ and $Pr(-sd(\boldsymbol{X},\mathcal{O}_i))$. 
Here we have a signed distance mapping $sd:\mathcal{X}\subset\mathbb{R}^2\rightarrow \mathcal{Y}\subset\mathbb{R}$.
Without loss of generality, we omit the obstacle parameter in $sd$ in the context of considering one static obstacle and define $Y^j=sd(\boldsymbol{X}^j,\mathcal{O}_i)=sd(\boldsymbol{X}^j)$ and $Y=sd(\boldsymbol{X},\mathcal{O}_i)=sd(\boldsymbol{X})$.
We then define the preimage of $sd$, i.e., $sd^{-1}:\mathcal{Y}\subset\mathbb{R}\rightarrow\mathcal{X}\subset\mathbb{R}^2$, such that $sd^{-1}(\mathcal{Y})=\{\boldsymbol{X}\in\mathcal{X}|sd(\boldsymbol{X})\in \mathcal{Y}\}$.
We derive the probability $Pr(Y\in\mathcal{Y})$, i.e.,

\begin{subequations}
\footnotesize
\begin{align}
Pr(Y\in\mathcal{Y})&=Pr(sd(\boldsymbol{X})\in \mathcal{Y})\label{eq:gmm_def}\\
&=\sum\limits_{q=1}^{Q} Pr(sd(\boldsymbol{X})\in \mathcal{Y}_q)\label{eq:gmm_def_monotonicity}\\
&=\sum\limits_{q=1}^{Q} Pr(\boldsymbol{X}\in sd^{-1}(\mathcal{Y}_q))\label{subeqn:prob_preimage_monotonicity}\\
&=Pr(\boldsymbol{X}\in sd^{-1}(\mathcal{Y}))\label{subeqn:prob_preimage}\\
&=\int_{\mathcal{X}} p(\boldsymbol{X}) d\boldsymbol{X}\label{subeqn:int_def}\\
&=\int_{\mathcal{X}} \sum_{j=1}^N p(\boldsymbol{X}|c(\boldsymbol{X})=j)p(c(\boldsymbol{X})=j) d\boldsymbol{X}\label{subeqn:gmm_def}\\
&=\sum_{j=1}^N \int_{\mathcal{X}} p(\boldsymbol{X}|c(\boldsymbol{X})=j)p(c(\boldsymbol{X})=j) d\boldsymbol{X}\\
&=\sum_{j=1}^N \int_{\mathcal{X}} p(\boldsymbol{X}|c(\boldsymbol{X})=j)\omega_j d\boldsymbol{X}\\
&=\sum_{j=1}^N \omega_j\int_{\mathcal{X}} p(\boldsymbol{X}|c(\boldsymbol{X})=j) d\boldsymbol{X}\label{equ:int of p(X_j)}\\
&=\sum_{j=1}^N \omega_j Pr(sd^{-1}(\mathcal{Y})|c(\boldsymbol{X})=j)\label{equ:sdf_Gaussian Component}\\
&=\sum_{j=1}^N \omega_j Pr(Y^j\in\mathcal{Y}).\label{equ:Prob of Y_j}
\end{align}
\end{subequations}

Due to the local monotonicity of $sd(\cdot)$, \cref{eq:gmm_def_monotonicity} can be derived into \cref{subeqn:prob_preimage_monotonicity} for each monotonic interval. 
As there exists a finite number of extrema in $sd(\cdot)$, \cref{eq:gmm_def} to \cref{subeqn:prob_preimage} can be established. 
The derivation to \cref{subeqn:int_def} holds because of the continuity of $sd(\cdot)$.
As $\mathcal{Y}$ is a Borel set, the $sd^{-1}(\mathcal{Y})$ is also measurable.
Derivation from \cref{subeqn:int_def} to \cref{subeqn:gmm_def} is based on the definition of GMM and Bayes' theorem, with $c(\boldsymbol{X})=j$ representing the event that $\boldsymbol{X}$ is sampled from the $jth$ GC. 
The reasoning from \cref{equ:int of p(X_j)} to \cref{equ:Prob of Y_j} involves the SDF defined under Gaussian uncertainty.

\end{proof}
\vspace{-0.4cm}

\subsection{Proof of \cref{pro:cvar-gmm expression}}
\label{appen:CVaR-GMM expression}
\begin{proof}
    According to the definition of CVaR in \cref{equ:cvar_min_Def} and the general properties of GMM, $CVaR_\alpha(Y)$ can be defined as 
\begin{subequations}
    \begin{align} 
    \label{eqn:cvar_def_L1}
CVaR_\alpha(Y)=&\min\limits_{z\in\mathbb{R}}\mathbb{E}\left[z+\frac{[Y-z]^+}{\alpha}\right]\\
=&\min\limits_{z\in\mathbb{R}}\left[z+\int_z^{+\infty}\frac{y-z}{\alpha} p(Y)dy\right]\\
=&\min\limits_{z\in\mathbb{R}}\left[z+\int_z^{+\infty}\frac{y-z}{\alpha} \sum\limits_{j=1}^N \omega_j p(Y^j) dy\right]\label{equ:gmmcvar_expression},
        \end{align}     
        \end{subequations} 
where $p(\cdot)$ denotes the PDF of a random variable.
For notation simplicity, we define
\begin{equation}
    Q(\boldsymbol{\omega},z)=z+\int_z^{+\infty}\frac{y-z}{\alpha} \sum\limits_{j=1}^N \omega_j p(Y^j) dy,
\end{equation}
where $\boldsymbol{\omega}=[\omega_1,\omega_2,\cdots,\omega_N]$, and thus we obtain
\begin{equation}
     CVaR_\alpha(Y)=\min\limits_{z\in\mathbb{R}} Q(\boldsymbol{\omega},z).\label{equ:cvar_Q}
\end{equation}
By taking the partial derivative of $Q(\boldsymbol{\omega},z)$ concerning $z$ as
\begin{align}
    \frac{\partial Q(\boldsymbol{\omega},z)}{\partial z}=1-\int_z^\infty\frac{p(Y)}{\alpha}dy,
    \end{align} 
     and leveraging the monotonicity of CDF, we can obtain the minimizer 
     \begin{equation}
    z^*=\argmin\limits_{z\in\mathbb{R}} Q(\boldsymbol{\omega},z)=VaR_\alpha(Y),\label{equ:unique minimizer}
     \end{equation} based on the definition of VaR. 
    
 Plugging $z^*$ into \cref{equ:gmmcvar_expression}, the $CVaR_\alpha(Y)$ can be further derived as follows:
 
\begin{subequations}
\small
\begin{align}
    &CVaR_\alpha(Y)\notag \\
    =&VaR_\alpha(Y) 
    +\int_{VaR_\alpha(Y)}^{+\infty} \frac{y-VaR_\alpha(Y)}{\alpha}\sum\limits_{j=1}^N \omega_j p(Y^j)dy\label{equ:cvar_exp}
    \\
    =&VaR_\alpha(Y) 
    +\frac{1}{\alpha}\sum\limits_{j=1}^N \omega_j \left[\int_{VaR_\alpha(Y)}^{+\infty} y p(Y^j) dy-\alpha_j VaR_\alpha(Y)\right]\label{equ:var_def}
    \\
    =&\frac{1}{\alpha}\sum\limits_{j=1}^N \omega_j \int_{VaR_\alpha(Y)}^{+\infty} y p(Y^j) dy \label{equ:var_equ}
    \\
    =&\frac{1}{\alpha}\sum\limits_{j=1}^N \omega_j \alpha_j CVaR_{\alpha_j}(Y^j).\label{equ:cvar_alphaj}
\end{align}
\end{subequations}
To obtain \cref{equ:var_def,equ:var_equ}, we define an auxiliary variable $\alpha_j$ as
\begin{subequations}
\begin{align}
&\alpha_j=\int_{VaR_\alpha(Y)}^{+\infty} p(Y^j) dy,\\
&\alpha=\sum\limits_{j=1}^N \omega_j \alpha_j,\label{equ:alpha_alphaj}
\end{align}
\end{subequations}
which refers to the tail probability of the $jth$ Gaussian SDF distribution at the $VaR_\alpha(Y)$ quantile.
The transformation to \cref{equ:cvar_alphaj} is based on the relationship of CVaR and VaR given in \cref{equ:cvar_min_Def}.
\end{proof}

\end{document}